\documentclass[letterpaper, 10 pt, conference]{ieeeconf}  

\IEEEoverridecommandlockouts                              
\usepackage{hyperref}
\usepackage{url}
\usepackage{graphicx} 
\usepackage{multirow}
\usepackage{caption}
\usepackage{subcaption}
\usepackage{amsmath}
\usepackage[utf8]{inputenc} 
\usepackage[T1]{fontenc}    
\usepackage{nameref}
\usepackage{caption} 
\usepackage{varioref}
\usepackage{hyperref}
\usepackage{cleveref}
\usepackage{float}
\usepackage{url}            
\usepackage{booktabs}       
\usepackage{amsfonts}       
\usepackage{nicefrac}       
\usepackage{microtype}      
\usepackage{lipsum}
\usepackage{float}
\usepackage{cite}
\usepackage{amsmath}
\usepackage{autobreak}

\usepackage{amsthm}
\usepackage[ruled,vlined]{algorithm2e}

\newtheorem{innercustomgeneric}{\customgenericname}
\providecommand{\customgenericname}{}
\newcommand{\newcustomtheorem}[2]{%
  \newenvironment{#1}[1]
  {%
   \renewcommand\customgenericname{#2}%
   \renewcommand\theinnercustomgeneric{##1}%
   \innercustomgeneric
  }
  {\endinnercustomgeneric}
}

\newcustomtheorem{customthm}{Theorem}
\newcustomtheorem{customlemma}{Lemma}
\newcustomtheorem{customcor}{Corollary}
\newcustomtheorem{customdef}{Definition}

\title{\LARGE \bf
Not Only Domain Randomization: Universal Policy with Embedding System Identification

}

\author{Zihan Ding
\\
 Princeton University
 \\
 zihand@princeton.edu
}

\begin{document}

\maketitle
\thispagestyle{empty}
\pagestyle{empty}

\begin{abstract}
Domain randomization (DR) cannot provide optimal policies for adapting the learning agent to the dynamics of the environment, although it can generalize sub-optimal policies to work in a transferred domain. In this paper, we present Universal Policy with Embedding System Identification (UPESI) as an implicit system identification (SI) approach with universal policies (UPs), as a learning-based control method to execute optimal actions adaptively in environments with various dynamic properties. Previous approaches of SI for adaptive policies either conduct explicit SI, which is testified to be an ill-posed problem, or suffer from low efficiency without leveraging the simulation oracle. We propose to conduct SI in the embedding space of system dynamics by leveraging a learned forward dynamics model, and use Bayesian optimization for the SI process given transition data in a new environment. The identified embeddings are applied as additional input to the UP to enable its dynamics adaptability. Experiments demonstrate the advantageous performances of our proposed UP with embedding SI over standard DR and conventional SI approaches on both low-dimensional and high-dimensional simulation tasks.

\end{abstract}


\section{INTRODUCTION}
Robot learning~\cite{yu2017preparing, peng2018sim, andrychowicz2020learning} has been testified to successfully work not only in simulation but also on real-world robotic control. In this domain, reinforcement learning (RL)~\cite{sutton1998introduction, duan2016benchmarking, dong2020deep} methods are typically applied for robotic control via sim-to-real transfer~\cite{peng2018sim, yu2019sim, valassakis2020crossing}. As one of the most widely applied methods in sim-to-real research literature, domain randomization (DR) can help to learn policies that are more general and robust to be applied in various environments which do not permit direct access to their underlying dynamics. System identification (SI) is another main category of methods that can help to bridge the domain gap. SI methods usually configure the physical parameters from historical transition data, either in an explicit or implicit manner. Both approaches have been proven to be feasible for some control tasks~\cite{andrychowicz2020learning, yu2017preparing}, when encountering the sim-to-real transfer problems, or more generally, domain transfer problems. However, a successful execution of the task does not necessarily indicate that an optimal control strategy is achieved, which leaves the space for further improvement based on solving the existing defects of above approaches. 

Problems of existing methods are investigated previously. Conventional DR methods~\cite{tobin2017domain} are able to provide a policy with better generalization in various environments, but not the optimal policy for the settings in a specific environment. The optimization objective is additionally taken with respect to an expectation of the distribution of system dynamic properties, rather than directly on the true dynamic settings of the testing environment. Therefore, being aware of the true dynamics of a new environment is necessary for accomplishing optimal control on a task, which is lacking for DR in present works. 

SI can help with bridging this gap via an active inferring of system dynamics with the necessary data collected.  In \cite{yu2017preparing}, an online system identification (OSI) module is used for explicitly configuring the dynamics parameters of environments, with a subsequent universal policy (UP) conditioned on the identified parameters to generate adaptive actions in new environments. However, both the UP and OSI modules can be infeasible in practice for general robot learning tasks according to the discussions in \cite{valassakis2020crossing}. OSI may not be feasible since the trajectories may not provide enough information for accurately predicting the dynamics parameters in a point-estimation manner, therefore distributions can better represent the uncertainty in the dynamics identification. The problem can be ill-posed to directly predict dynamics parameters, according to \cite{zhu2018efficient}. Multiple potential dynamics models can lead to the same trajectories as a non-injective function, due to the entanglement of dynamics parameters. For UP, it can be hard to train when the dimension of dynamics parameters is high.

Present methods for system dynamics identification usually require aligned trajectories for constructing the contrastive loss~\cite{zhu2018efficient, chebotar2019closing}, which we find not to be a necessity. In previous methods, the trajectories alignment can usually be achieved with setting same initial states and executing the same policy, or rollout the simulation environment on a give state and certain action for a single timestep at each time. However, not all simulators can support a direct setup to be any state in the environments. Setting up the initial states only will lead to compounding errors on the trajectories due to the dynamics difference or different random seeds, or requires undesirable engineering efforts. Our proposed method solves the trajectories alignment procedure by leveraging on a well-trained forward dynamics prediction model.

This paper makes the following contributions:
1) We demonstrate the limited performances of typical DR methods when the dynamics of testing environment is uncertain or various, compared against our proposed method; 
2) We propose a method to handle dynamics uncertainty and unmodeled effects~\cite{andrychowicz2020learning} separately, with SI module for implicitly identifying the dynamics parameters, as well as DR for unmodeled effects involving observation noise, observation delay and action noise;
3) For system dynamics identification, our method does not depend on aligned trajectories collected in the source and target domains, but with randomly sampled trajectories carried out by the same control policy;
4) Our method learns a regularized dynamics embedding rather than applying the oracle dynamics parameters as the representation of system dynamics, to handle the entangled effects or redundant information within dynamics parameters. A universal policy conditioned on the learned dynamics embedding is trained with DR for both dynamics and unmodeled effects. 

\section{RELATED WORK}
Existing methods for bridging the domain transfer gap (\emph{e.g.}, sim-to-real) can be broadly categorized in to the following classes: 1). Domain Randomization~\cite{tobin2017domain} randomizes either visual features~\cite{james2017transferring} or dynamics parameters~\cite{peng2018sim} in a source domain to train policies with better adaptability to the target domain. The branch of works also include the recent process of structured DR~\cite{prakash2019structured} and active DR~\cite{mehta2020active}. 2). System Identification usually requires an extensive data collection and calibration process to mitigate the gap between the source and target environments. Explicit SI can be incorporated with a universal policy~\cite{yu2017preparing} for achieving an adaptive control of the real robot for various environment settings. Implicit SI usually uses recurrent units like long short-term memory (LSTM) networks with sequential inputs to preserves information about environment dynamics~\cite{peng2018sim, andrychowicz2020learning}. 3). Domain Adaptation (DA)~\cite{wang2018deep, tanwani2020domain} applies transfer learning techniques to match the distribution of source domain data with the target domain data, often applied in visuomotor control with images.  4). Strategy Optimization (SO)~\cite{yu2018policy, yu2019sim, yu2020learning} requires evaluating a family of policies (called \textit{strategy}) in the target domain and selecting the one with the best performance, usually with sampling-based methods like Bayesian optimization (BO)~\cite{mockus2012bayesian} and CMA-ES~\cite{hansen1995adaptation}. 

Our works focus on the SI approach with universal policies trained for domain transfer. DR in general cannot provide optimal policy due to the induced noise for increasing model generality. The optimized objective for the control policy is additionally taken with respect to an expectation of randomized parameters in environments. Thus only sub-optimal policies can be performed in the target domain even for the case that target sample data falls in the domain of randomized distributions of the source. Traditional explicit SI methods~\cite{yu2017preparing} directly configure the system parameter, which is testified to be an ill-posed problem~\cite{zhu2018efficient,valassakis2020crossing} due to that the entangled effects of multiple parameters will lead to non-unique identification results. This problem no longer exists for implicit SI~\cite{peng2018sim, andrychowicz2020learning}. However, the true system parameters in the source domain are not leveraged for implicit SI, which hinders the learning efficiency of this approach. To this end, we found the embedding of system dynamics parameters can be achieved with encoders to improve the above approaches. 

\section{PRELIMINARIES}
\subsection{Notations}
A typical formulation of RL problems follows a standard Markov decision process (MDP), which can be represented as $(\mathcal{S}, \mathcal{A}, R, \mathcal{T}, \gamma)$, where $\mathcal{S}$ and $\mathcal{A}$ are feasible sets of state and action, and $R$ is reward function $R(s,a)$: $\mathcal{S}\times \mathcal{A}\rightarrow \mathbb{R}$. $\mathcal{T}$ defines the transition probability from current state $s$ and action $a$ to a next state $s^\prime$ based on a \textbf{fixed} dynamics setting: $\mathcal{T}(s^\prime|s,a)$, and $\gamma\in(0,1)$ is a discount factor. However, it is not sufficient for a learned policy to be applied in a transferred domain. We consider a partially observable MDP (POMDP) with \textbf{randomized} dynamics as $(\mathcal{S}, \mathcal{A}, \Theta, \mathcal{T}, \mathcal{O}, \mathcal{P}_{\mathcal{O}}, R, \gamma)$, with additional $\Theta$ as the $\mathbb{R}^d$ space of dynamics parameters ($d$ is the number of dynamics parameters), $\mathcal{O}$ as the observation space and $\mathcal{P}_{\mathcal{O}}$ as the emission probability distribution: $\mathcal{P}_{\mathcal{O}}(o|s)$. Moreover, the transition probability $\mathcal{T}$ will then be different from the standard MDP, and it further depends on the dynamics parameters $\theta$ as $\mathcal{T}(s^\prime|s,a,\theta), \theta\in\Theta$. Note that here we define $\mathcal{P}_{\mathcal{O}}$ to be independent on dynamics parameters $\theta$, since we disentangle the observed transition process into the transition $\mathcal{T}$ in underlying state space $\mathcal{S}$ and the omission $\mathcal{P}_{\mathcal{O}}$ of observations from states, so only the transition of underlying states $\mathcal{T}$ depends on the dynamics parameters $\theta$.

\section{METHODOLOGY}
\begin{figure*}[htbp]
    \begin{center}
        \includegraphics[scale=0.26]{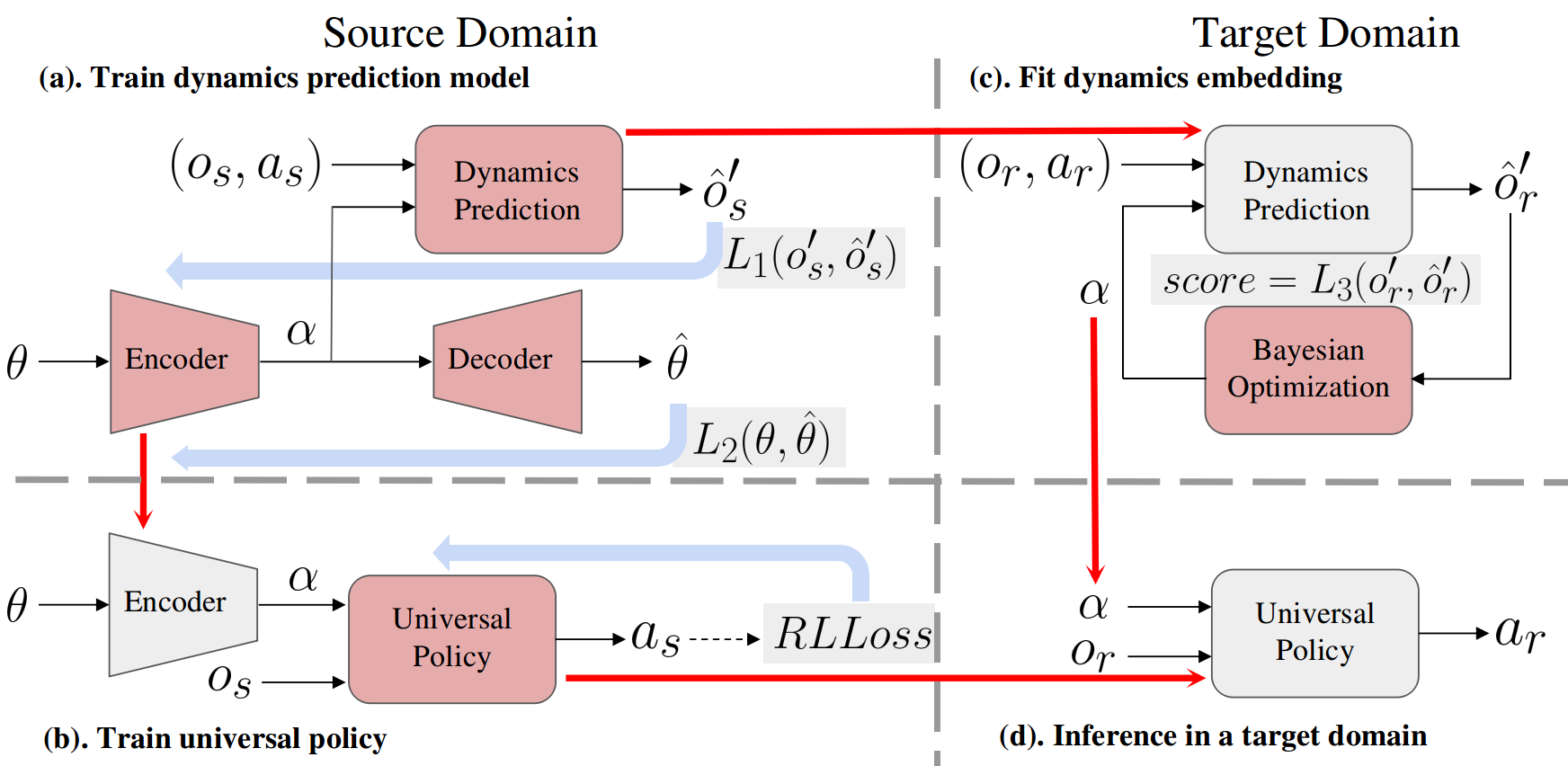}
    \end{center}
    \caption{Overview of our method, including four steps: (a) training the dynamics prediction model with inputs of observation $o$, action $a$, and embedding $\alpha$ of system dynamics parameter $\theta$; (b) training the universal policy with a learned encoder for providing the embedding of system dynamics; (c) optimization in the embedding space with data from a target domain using Bayesian optimization and the learned forward prediction model; (d) inference in the target domain environment with identified dynamics embedding and the universal policy. The blue shaded arrows indicate the gradient flows during model training. The red arrows indicate a trained/fitted models or parameters are applied (with fixed weights/values) somewhere else. }
    \label{fig:overview}
\end{figure*}
In this paper, we propose the method for domain transfer with not only DR but also SI in an embedding space. The overview of our method is shown in Fig.~\ref{fig:overview}. Three key components are involved in the framework:
 \begin{center} 
 \textbf{Dynamics Encoder}: $\mathbf{E}(\theta) \mapsto \alpha$  \\
 \textbf{Forward Dynamics Predictor}: $\mathbf{F}(o,a,\alpha)\mapsto o^\prime$ \\
 \textbf{Universal Policy}: $\pi(o, \alpha) \mapsto a$ \\
 \end{center}
 
The \textbf{Dynamics Encoder} $\mathbf{E}$ is an embedding network to generate low-dimensional embedding $\alpha$ from the system parameters $\theta$. In practical robotic learning tasks, dozens of system parameters can be involved in simulation~\cite{valassakis2020crossing, andrychowicz2020learning}, which would severely increase the difficulty of universal policy learning. Moreover, explicit identification of each parameter value is neither necessary nor impossible for general cases~\cite{zhu2018efficient}. For example, an increment of friction coefficients on a specific joint may be counteracted by a decrease of mass of some link bodies when we simply look at the trajectories executed by the robot with the same torques applied. The entangled effects of system parameters are considerable in SI process. The \textbf{Forward Dynamics Predictor} $\mathbf{F}$ is learned to mirror both the transition function $\mathcal{T}(s^\prime|s,a,\theta)$ of the environment with randomized dynamics $\theta\in\Theta$ and the observation emission function $\mathcal{P}_{\mathcal{O}}(o|s)$ for transferring states $s,s^\prime$ to observations. By leveraging the conditional dependency on embedding $\alpha$ in function $\mathbf{F}$, we propose to directly optimize $\alpha$ in the embedding space using Bayesian optimization when given the data from the target domain, for achieving the embedding SI. The \textbf{Universal Policy} $\pi(o, \alpha)$ basically follows the standard settings in \cite{yu2017preparing}, where a concatenation of observations and identified parameter embedding is taken as policy inputs. By conditioning the policy on the system properties, it is capable of adaptively selecting optimal actions in different system settings and randomized parameters $\theta$. We will show details in later sections.

\subsection{Theoretical Insights}

In the following, we provide the theoretical justification of our method. Since we assume the observation emission process has no impact on the dynamics of underlying state space, the analysis below will be based on a fully observable MDP with randomized dynamics but no observations. Apart from this, another difference of our implementation in practice from the theoretical analysis is the usage of dynamics embedding, \emph{i.e.}, using $\alpha$ instead of $\theta$. These two differences will not cause the loss of generality for our analysis.

Following a similar setup in \cite{yang2019imitation}, we extend the definitions of different occupancy measures for MDP with randomized dynamics as in Table~\ref{tab:occupancy}.

\begin{center}
\begin{table*}[htbp]
\captionof{table}{Different occupancy measure for MDP with randomized dynamics}
\begin{tabular}{ |c|c|c|c|c| } 
\hline
Occupancy Measure & State-Action  & State Transitions & Joint w/o Dynamics & Joint w/ Dynamics \\
\hline
Denotation & $\rho_\pi(s,a, \theta)$ &  $\rho_\pi(s,s^\prime)$ & $\rho_\pi(s,a,s^\prime)$ & $\rho_\pi(s,a,\theta,s^\prime)$  \\ \hline
Support & $\mathcal{S}\times\mathcal{A}\times\Theta$ & $\mathcal{S}\times\mathcal{S}$ & $\mathcal{S}\times\mathcal{A}\times\mathcal{S}$ & $\mathcal{S}\times\mathcal{A}\times\Theta\times\mathcal{S}$  \\  \hline
Definition & $\rho_\pi(s,\theta)\pi(a|s,\theta)$ & $\int_{\mathcal{A}\times\Theta}\rho_\pi(s,a,\theta)\mathcal{T}(s^\prime|s,a,\theta)\text{d} a \text{d}\theta$ &
$\int_{\Theta}\rho_\pi(s,a,\theta)\mathcal{T}(s^\prime|s,a,\theta)\text{d}\theta$ & $\rho_\pi(s,a,\theta)\mathcal{T}(s^\prime|s,a,\theta)$  \\ \hline
\end{tabular}
\label{tab:occupancy}
\end{table*}
\end{center}

\begin{customdef}{1}[Estimated Dynamics]
The estimated dynamics distribution is defined as the distribution of dynamic parameters estimated from given the transition tuples $\{(s,a,s')\}$, and it can be calculated as:
\begin{equation}
    p(\theta|s,a,s') = \frac{\rho(s,a,\theta,s')}{\rho(s,a,s')} = \frac{\mathcal{T}(s'|s,a,\theta)\rho(s,a,\theta)}{\int_{\Theta}\mathcal{T}(s'|s,a,\bar{\theta})\rho(s,a,\bar{\theta})\text{d}\bar{\theta}}
\end{equation}
\end{customdef}
In practice, the estimated dynamics can be represented as a parameterized model with learnable parameters $\psi$ and optimized with dataset $\mathcal{D}^s$ from the source domain. During training or evaluation, the predicted distribution and true distribution are denoted as $p^s_\psi(\theta|s,a,s')$ and $p^t(\theta|s,a,s')$ respectively. The optimization objective for SI can be formulated as minimizing the discrepancy of estimated and true dynamics distributions, \emph{e.g.}, $\min_\psi \mathbb{D}_\text{KL}[p^s_\psi(\theta|s,a,s')||p^t(\theta|s,a,s')]$. We will prove that the method of leveraging the dynamics prediction model to achieve SI (step (c). in the scheme as in Fig.~\ref{fig:overview}) is a valid approach for this objective.

\begin{customlemma}{2}
Given the same dataset distribution $\rho^s(s,a,s')=\rho^t(s,a,s')$, the difference between the estimated dynamics distribution $p^s$ and the true distribution $p^t$ can be characterized as:
\begin{align}
&\mathbb{D}_\text{KL}[p^s(\theta|s,a,s')||p^t(\theta|s,a,s')] \\
    &=\mathbb{D}_\text{KL}[\rho^s(s,a,\theta,s')||\rho^t(s,a,\theta,s')]
\end{align}
\label{the:1}
\end{customlemma}

\begin{customdef}{3}[Forward Dynamics Prediction]
The forward dynamics prediction model is defined as the distribution of next state $s'$ given the current tuple $(s,a,\theta)$ in a MDP with randomized dynamics, as an approximation of the true underlying dynamics in the source/target domain:
\begin{equation}
    f(s'|s,a,\theta)\approx \mathcal{T}(s'|s,a,\theta)=\frac{\rho(s,a,\theta,s')}{\rho(s,a,\theta)}
\end{equation}
\end{customdef}

In practice, this model approximation process can be achieved with optimization on collected dataset in the source domain. We denote the parameters of forward dynamics prediction as $\Phi$, which are learned in the approximation process with objective $\min_\Phi \mathbb{D}_\text{KL}[f^s_\Phi(s'|s,a, \theta)||f^t(s'|s,a, \theta)]$ (step (a).), where $\{(s,a,\theta)\}$ are given and $f^t(s'|s,a, \theta)$ are the true distributions of the forward prediction. After training, $f(s'|s,a,\theta)$ is evaluated on the testing dataste (\emph{e.g.}, the target domain dataset), with its predicted distribution denoted as $f^s_\Phi(s'|s,a,\theta)\approx\frac{\rho^s(s,a,\theta,s')}{\rho^s(s,a,\theta)}$ and true distribution as $f^t(s'|s,a,\theta)\approx\frac{\rho^t(s,a,\theta,s')}{\rho^t(s,a,\theta)}$. In our proposed method leveraging the forward dynamics prediction, the SI process is characterized as $\min_\theta \mathbb{D}_\text{KL}[f^s_\Phi(s'|s,a, \theta)||f^t(s'|s,a, \theta)]$ (step (c).), where $\{(s,a)\}$ and $\Phi$ are given. In conventional SI, without $f(s'|s,a,\theta)$, this optimization objective actually becomes $\mathbb{D}_\text{KL}[\mathcal{T}^s(s'|s,a, \theta)||\mathcal{T}^t(s'|s,a, \theta)]$, where $\mathcal{T}^s$ and $\mathcal{T}^t$ are the transition functions of the source and target domains respectively, and it requires the simulator to set the same $(s,a)$ as the real world does all the time for a sim-to-real transfer example.

Now we have another lemma of discrepancy between $f^s(s'|s,a,\theta)$ and $f^t(s'|s,a,\theta)$ as follows.
\begin{customlemma}{4} The distance of the distribution from the forward dynamics prediction and the true distribution can be formulated with the KL-divergence, it thus follows:
\begin{align}

\mathbb{D}_\text{KL}[f^s||f^t]\approx \mathbb{D}_\text{KL}[\rho^s||\rho^t] -\mathbb{D}_\text{KL}[\rho^s(s,a,\theta)||\rho^t(s,a,\theta)]
\end{align}
\label{the:2}
where $f^{s,t}$ are shorten for $f^{s,t}(s'|s,a, \theta)$, and $\rho^{s,t}$ are shorten for $\rho^{s,t}(s,a,\theta,s')$.
\end{customlemma}

\begin{customthm}{5}
The optimization of forward dynamics prediction is increasing the lower bound of the optimization objective for improving the estimated dynamics, i.e.,

\begin{equation}
    \mathbb{D}_\text{KL}[p^s||p^t]\geq  \mathbb{D}_\text{KL}[f^s||f^t]
\end{equation}
where $p^{s,t}$ are $p^{s,t}(\theta|s,a,s')$ and $f^{s,t}$ are $f^{s,t}(s'|s,a,\theta)$.
\label{thm:lb}
\end{customthm}
The above theorem tells us that a direct optimization of estimated dynamics for parameters $\phi$ can be approximately achieved with optimization of parameters $\Phi$ for forward dynamics prediction then leverage it to optimize $\theta$. 
Proofs of above theorems and lemmas are all provided in App.~\ref{appsec:proof}.

In further analysis, we note that step (a) and (c) in our method are actually a SI process, not in original parameter space but an embedding space. As proposed in \cite{tanwani2020domain}, the domain-invariant representation learning can be achieved by both minimizing the marginal discrepancy (called \textit{marginal distributions alignment}) and minimizing the conditional discrepancy (called \textit{conditional distributions alignment}) for an object classification task. Here we try to achieve a domain-invariant forward prediction model by optimizing the embedding of system dynamics. We will show how our method satisfies the process of minimizing conditional discrepancy. The reason that marginal distribution alignment is not achieved in current settings is due to the lack of true system dynamics parameters in the target domain, which are required in marginal distribution alignment.

For conditional distributions alignment, we first modify the definition in \cite{tanwani2020domain} with additional conditional variable. Then we narrow down the definition for general domain-transfer models to the specific forward dynamics prediction model in our method.
\begin{customdef}{6}[Conditional Distributions Alignment (modified from \cite{tanwani2020domain})]
Given two domains $D^s=\{x^s_i, y^s_i, z^s_i\}^{N^s}_{i=1}$ and $D^T=\{x^t_i, y^t_i, z^t_i\}^{N^t}_{i=1}$ drawn from random variables $(X^s\times Y^s\times Z^s)$ and $(X^t\times Y^t\times Z^t)$ with different output conditional probability distributions $Pr(Y^s|X^s, Z^s)\neq Pr(Y^t|X^t, Z^t)$, conditional alignment corresponds to finding the transformation $X\times Z \overset{g(X)}{\longrightarrow} E\times Z \overset{f(E,Z)}{\longrightarrow} Y$ such that the discrepancy between the transformed conditional distributions is minimized, \emph{i.e.}, $Pr(Y^s|g(X^s), Z^s)=Pr(Y^t|g(X^t), Z^t)$. Note that the number of additional variables $Z$ is not limited to be one.

\end{customdef}
In our case with forward dynamics prediction model $\mathbf{F}(o,a,\alpha)$ and dynamics encoder $\mathbf{E}(\theta)$, we have $Pr(o^{t \prime}|o^t, a^t, \theta^t)\neq Pr(o^{s \prime}|o^s, a^s, \theta^s)$, and conditional alignment for the forward dynamics prediction model optimizes the dynamics embedding $\alpha$ such that, based on the optimized dynamics embedding, the resulting prediction distributions are minimized, \emph{i.e.}, $Pr(o^{s \prime}|o^s,a^s;\mathbf{E}(\theta^s))=Pr(o^{t \prime}|o^t,a^t;\alpha^t)$. In the source domain the embedding $\alpha^s=\mathbf{E}(\theta^s)$ is optimized via back-propagating through the dynamics encoder $\mathbf{E}$, while in the target domain $\alpha^t$ is directly optimized with methods like BO due to the lack of true dynamics parameters. From above analysis, we show that our method (specifically step (a) and (c)) accomplishes an extended conditional distribution alignment for identifying the dynamics embedding.

\subsection{Universal Policy with Embedding System Identification}
In previous methods, SI is usually achieved with aligned trajectories~\cite{zhu2018efficient, chebotar2019closing}, which is found to be unnecessary in our method. We will detail the formulation of two approaches as follows. Suppose the datasets in the source and target domains are as $\mathcal{D}^s=\{(o^s, a^s, o^s{}^\prime)\}$ and $\mathcal{D}^t=\{(o^t, a^t, o^t{}^\prime)\}$, $\mathcal{D}^s$ is collected with dynamics randomization (without observation noise and action noise). Apart from deriving actions from the same policy as $a^s=\pi(o^s)$ and $a^t=\pi(o^t)$, most present methods assume the underlying states to be the same in the source and target domains, as $o^t\sim\mathcal{P_O}(\cdot|s)$ and $o^s\sim\mathcal{P_O}(\cdot|s)$ for $\forall o^t\in \mathcal{D}^t, o^s\in\mathcal{D}^s$, so as to define a loss function as $L(\{o^s{}^\prime, o^t{}^\prime\})$  for optimizing the identified system dynamics parameters. The trajectory in underlying state space $\tau=(s_{T=0}, s_{T=1}, s_{T=2}, ..., s_{T=n})$ is required to be aligned in the source domain and the target domain to form a valid loss function, especially for the initial state $s^{T=0}$. In our experiments, we find this assumption in dataset collection both unnecessary and inconvenient. On the one hand, due to the potential dynamics differences, the underlying states may not be well aligned for the dataset collected in source and target, \emph{e.g.}, $\tau^s=(s_{T=0}^s, s_{T=1}^s, s_{T=2}^s, ..., s_{T=n}^s)\neq \tau^t=(s_{T=0}^t, s_{T=1}^t, s_{T=2}^t, ..., s_{T=n}^t)$ especially for the latest states in trajectories. On the other hand, manually set the states for the source domain to match with the target domain may not be feasible or inconvenient for some practical cases, \emph{e.g.}, hard to set a simulator to a certain state, etc. Therefore, in our proposed method, the trajectory alignment is no longer required for the SI process. Specifically, we assume that a forward dynamics prediction model $\mathbf{F}$: $\hat{o^\prime}=\mathbf{F}(o,a;\theta)$ trained with source domain dataset $\mathcal{D}^s$ will accurately predict the next observations in the target domain dataset $\mathcal{D}^t$ only when the input dynamics embedding $\alpha$ of the target environment is accurate. Therefore we can construct a loss/score function for optimizing the dynamics embedding based on the forward prediction results in the target domain.

In our method, the forward dynamics prediction model $\mathbf{F}$: $\hat{o}^\prime=\mathbf{F}(o,a;\alpha)$, is trained with $\mathcal{D}^s$ and further applied on $\mathcal{D}^t$ to fit $\alpha$, assuming that the target data is within the distribution of source data. This can be achieved by increasing the randomization ranges in the source domain until satisfactory. For learning the forward prediction model $\mathbf{F}_\Phi$ and dynamics encoder $\mathbf{E}_\psi$, we have the following objectives:
\begin{align}
    &\min_\Phi \mathbb{E}_{(o,a,o^\prime, \theta)\sim\mathcal{D}^s}[L_1(\mathbf{F}_\Phi(o,a;\mathbf{E}(\theta)), o^\prime)] \label{eq:forward_prediction} \\
    &\min_\psi \mathbb{E}_{(o,a,o^\prime, \theta)\sim\mathcal{D}^s}[L_1(\mathbf{F}(o,a;\mathbf{E}_\psi(\theta)), o^\prime)+\lambda L_2(\mathbf{D}(\mathbf{E}_\psi(\theta)))] \label{eq:dynamics_encoder}
\end{align}
where both $L_1$ and $L_2$ are mean squared error (MSE) loss function in our experiments, and $\lambda$ is a trade-off coefficient for balancing the dynamics prediction performance and the reconstruction of dynamics parameters through encoding and decoding. This is corresponding to step (a) in Fig.~\ref{fig:overview}.

The objective for optimizing the embedding $\alpha$ with the learned forward dynamics prediction function $\mathbf{F}_{\Phi^*}$ is:
\begin{equation}
    \alpha^* = \arg \min_\alpha \mathbb{E}_{(o,a,o^\prime)\sim\mathcal{D}^t}[L_3(\mathbf{F}_{\Phi^*}(o,a;\alpha), o^\prime)]
    \label{equ:embed}
\end{equation}
where $L_3$ is also MSE in our experiments. This is corresponding to step (c) in Fig.~\ref{fig:overview}.

The universal policy $\pi$: $a=\pi(o;\alpha)$ in our method is trained in the source domain and applied for inference in the target domain. During the training process of the universal policy, observation noise and action noise are applied with randomized parameters, together with the randomization of dynamics parameters. The objective for learning the universal policy is:
\begin{equation}
    \phi^* = \arg \max_\phi \mathbb{E}_{\theta}[J_\theta(\pi_\phi(\cdot|\mathbf{E}(\theta)))]
    \label{equ:up}
\end{equation}
where $\phi$ are parameters of the policy $\pi$. This is corresponding to step (b) in Fig.~\ref{fig:overview}.

Combining above objectives as in Eq.~\eqref{equ:embed} and \eqref{equ:up}, our method is capable of achieving the optimal policy inference for any dynamics setting via optimized embedding SI:
\begin{equation}
    \pi^* = \pi_{\phi^*}(\cdot|\alpha^*)
\end{equation}
which corresponds to step (d) in Fig.~\ref{fig:overview}. The pseudo-code for the entire UPESI algorithm is shown as Alg.~\ref{alg:upesi}.

\begin{algorithm}[htbp]
\SetAlgoLined
 \# (a). TRAIN DYNAMICS PREDICTION MODEL\;
 Initialize data buffer $\mathcal{D}^s$, prediction model $\mathbf{F}_\Phi$ and encoder $\mathbf{E}_\psi$, and a pretrained policy $\pi_\text{ini}$\;
 \While{$i<N_1$}{
    $\theta\in\Theta$\;
    \While{$j<N_2$}{
    Execute policy $\pi_\text{ini}$ on robot to collect samples $(o,a,o^\prime)$ under dynamics $\theta$;
    $\mathcal{D}^s=\mathcal{D}^s\bigcup (o,a,o^\prime, \theta)$\;
    }
 }
 \While{$i<N_3$}{
    Sample a batch of data $\mathcal{D}^b$ from $\mathcal{D}^s$\;
    Optimize $\mathbf{F}_\Phi$ and $\mathbf{E}_\psi$ with Eq.~\eqref{eq:forward_prediction} and Eq.~\eqref{eq:dynamics_encoder} respectively using $\mathcal{D}^b$ \;
    }
 
 \# (b). TRAIN UNIVERSAL POLICY\;
    Fix encoder parameters $\mathbf{E}_\psi$\;
    Initialize universal policy $\pi_\phi$\;
    \While{$i<M$}{
    $\theta\in\Theta$\;
    $\alpha=\mathbf{E}_\psi(\theta)$\;
    \While{$j<\text{MaxSteps}$}{
    $a=\pi_\phi(o;\alpha)$\;
    $o^\prime, r$=env.step($a$;$\theta$)\;
    Update $\pi_\phi$ with Eq.~\eqref{equ:up}\;
    }
    }
 
 \# (c). FIT DYNAMICS EMBEDDING\;
    Fix dynamics prediction model $\mathbf{F}_\Phi$\;
    Collect data buffer $\mathcal{D}^t$ in the target domain following the same way as the dynamics training data buffer $\mathcal{D}^s$\;
    \While{$i<M$}{
    Bayesian optimize the embedding $\alpha$ with $\mathcal{D}^t$ as Eq.~\eqref{equ:embed};
    }

 \# (d). INFERENCE IN A TARGET DOMAIN\;
    Rollout the universal policy $\pi$ to generate action: $a=\pi(o;\alpha)$.
 \caption{Universal Policy with Embedding System Identification (UPESI)}
 \label{alg:upesi}
\end{algorithm}




\section{EXPERIMENT}
\subsection{Comparison Methods}
The comparison involves the following methods:

(1). \textbf{No DR.} A conservative policy is trained in a certain environment with DR, as a comparison baseline. 

(2). \textbf{DR only.} A general policy is trained with randomized environments on specified parameters and distributions.

(3). \textbf{DR+UP (True).} An adaptive universal policy is trained with true system parameters as additional inputs in randomized environments.

(4). \textbf{DR+UP+SI.} The policy is trained in the same manner as (3), but tested with a learned SI module to predict dynamics parameters from historical transitions.

(5). \textbf{DR+UP+Encoding (BO).} As the proposed method, an adaptive universal policy is trained with the embedding of system parameters as additional inputs in randomized environments, also with a BO process for configuring the embedding in new environment using the learned forward prediction model.

(6). \textbf{DR+UP+Encoding (True).} As an oracle for the proposed method, the universal policy is trained in the same way as (5), but the embedding is given by the true system parameters going through the learned encoder.



Method (1) works as a baseline for all other methods, which basically represents the optimal policy for a certain training environment and also the most conservative policy when testing in various environments. The comparison of (3) and (4) will imply the potential effects caused by the deficiency in SI module. The comparison of (5) and (6) will show the effects of embedding configuration based on BO when there are no true system parameters but only with samples of transition provided.

\subsection{Experimental Setup}
\begin{figure}[htbp]
	\centering\includegraphics[width=8.5 cm]{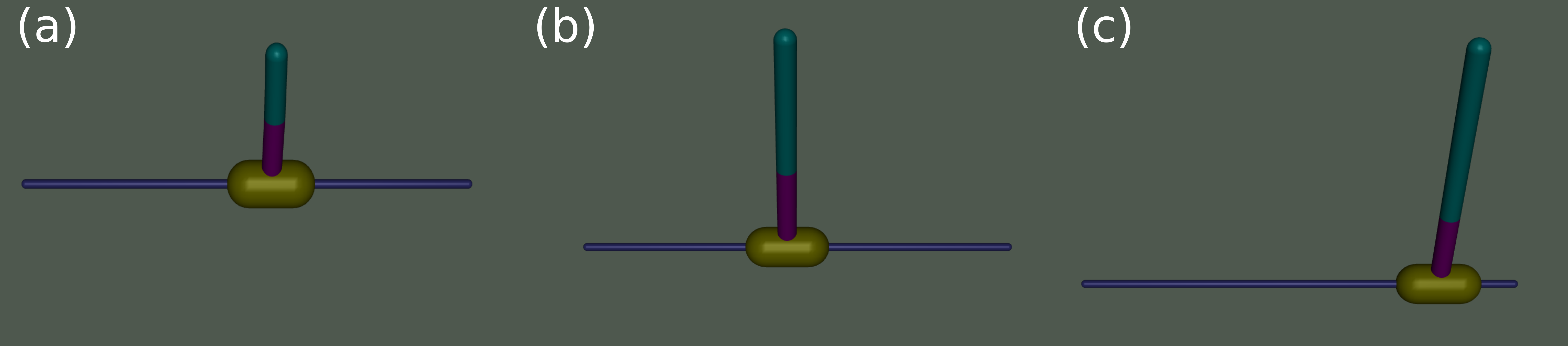}
	\caption{Different configurations for the environment \textit{InvertedDoublePendulum-v2}.}
	\label{fig:different_configs_pendulum}
\end{figure}
Environments \textit{InvertedDoublePendulum-v2}, \textit{HalfCheetah-v3} in OpenAI Gym MuJoCo are used for testing  the above methods.
For \textit{InvertedDoublePendulum-v2}, the environment is randomized for its five important dynamics parameters, including damping, gravity, two geometry lengths and density, which are detailed in App.~\ref{appsec:rand} Tab.~\ref{tab:pendulum}. The encoded latent space is chosen to have a dimension of 2 as the disentangled compact representation of system dynamics. Fig.~\ref{fig:different_configs_pendulum} shows visualization of three different configurations for the task scene with randomized parameters. For \textit{Halfcheetah-v3}, 13 dynamics parameters are randomized as detailed in App.~\ref{appsec:rand} Tab.~\ref{tab:halfcheetah}, with an embedding space of dimension 4. For training both the conservative or adaptive universal policies, we use the twin delayed deep deterministic policy gradient (TD3) algorithm~\cite{fujimoto2018addressing}, with 4 MLP layers for both the policy and the value networks. The dynamics networks and SI networks have 4 MLP layers. SI model predicts the system dynamics parameters with a stack of 5 frames of transitions, using the same training data as in the training of dynamics prediction models. In our experiments, each method is tested with three runs with different random seeds.

\subsection{Experimental Results}
\begin{figure}[htbp]
	\centering\includegraphics[width=9 cm]{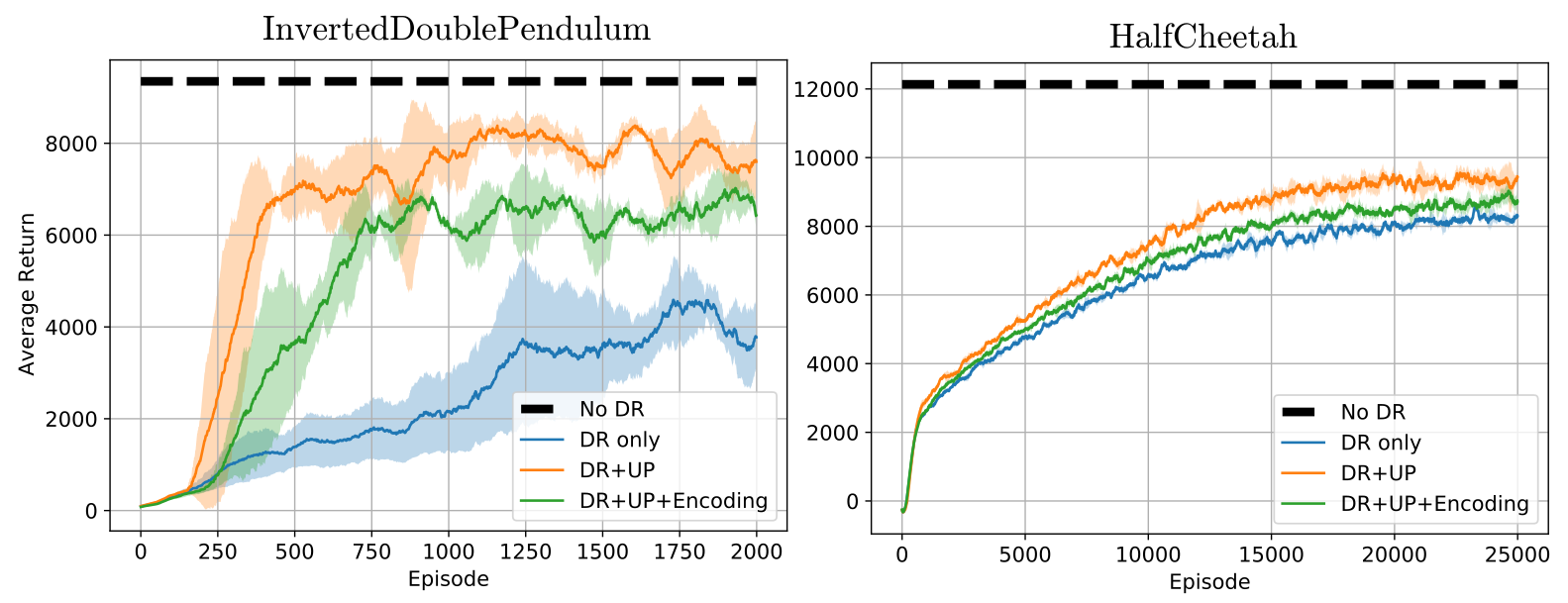}
	\caption{Average learning performances of different methods for two environments. Methods include no DR (black dashed lines), DR only, DR with UP, DR with UP and encoding. }
	\label{fig:learning_curve}
\end{figure}

Fig.~\ref{fig:learning_curve} shows the average learning performances in terms of episode rewards in training for all methods on two environments. The policies trained without DR are evaluated in their specific training environments as optimal baselines. The shaded areas indicating the standard deviations. Each run takes 2000 or 25000 episodes of training the policies with TD3 algorithm for \textit{InvertedDoublePendulum-v2} and \textit{HalfCheetah-v3} respectively. The universal policies with either embedding SI or true system parameters significantly outperforms the DR baselines in two environments, although not as optimal as the no-DR policies since the episode reward is directly evaluated in the training environments (\emph{i.e.} non-randomized for no-DR case but randomized for others).

\begin{figure}[htbp]
	\centering\includegraphics[width=9 cm]{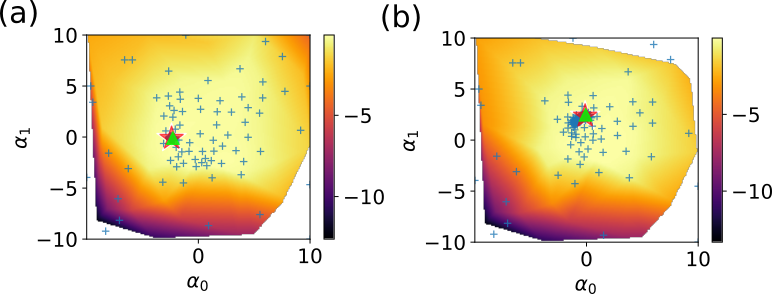}
	\caption{Bayesian optimization on two-dimensional dynamics embedding space for the environment \textit{InvertedDoublePendulum-v2}. Two cases with different dynamics parameters are displayed. The blue "crosses" are the query points in BO process, the red "stars" are the best points after BO, and the green "triangles" are the true values for dynamics embedding derived from the true system parameters. Note that the ``triangles" and ``stars" are almost overlapped.}
    \label{fig:bo_pendulum}
\end{figure}
For the embedding SI process, we take the \textit{InvertedDoublePendulum-v2} as an example. Fig.~\ref{fig:bo_pendulum} shows the BO process for configuring the embeddings of two sets of randomly sampled system parameters following the step (c) in our proposed method (as in Fig.~\ref{fig:overview}), with the dynamics prediction model trained with 10000 episodes of policy rollouts and BO for each parameter configuration with 1000 episodes of data. After 500 iterations of BO for embedding configuration, the best query point is already very close to the true embedding values as shown in Fig.~\ref{fig:bo_pendulum}. Results are similar for \textit{HalfCheetah-v3} just with a higher dimensional embedding. 2000 episodes of data are leveraged for training the dynamics prediction model and 100 episodes of data are used for BO in identifying each new testing environment.
\begin{figure}[htbp]
	\centering\includegraphics[width=8 cm]{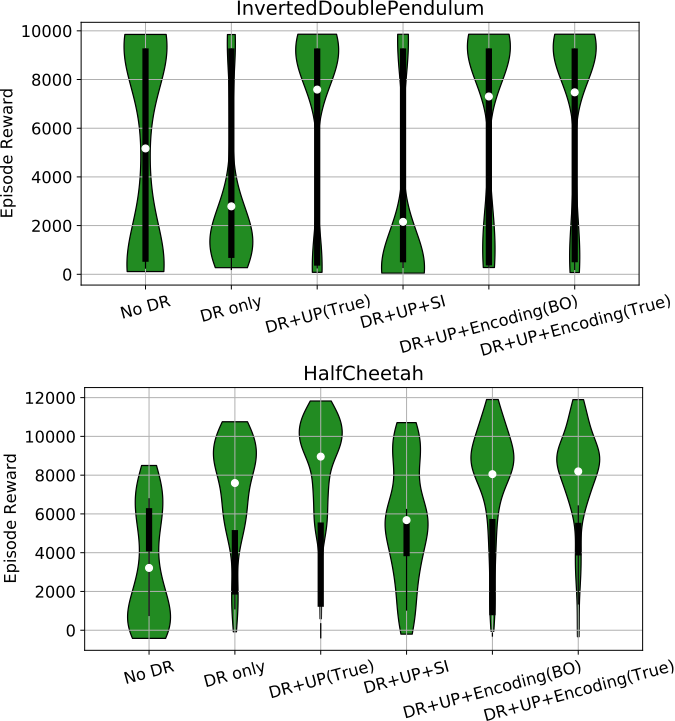}
	\caption{Test performances of different methods for two environments. The white dots are the mean values, and the black bold lines indicate the 25\%-75\% quantiles.}
	\label{fig:compare_methods}
\end{figure}

\begin{table}[htbp]
\begin{center}
\captionof{table}{Test performances: mean and standard deviations of episode rewards for different methods on two environments.}
\begin{tabular}{ |c|c|c|c| } 
\hline
Env & Method &  Episode Reward \\
\hline
\multirow{6}{*}{\textit{pendulum}} & No DR & $5168.9\pm{4324.4}$  \\ \cline{2-3}
 & DR only & $2791.1\pm{2960.8}$\\ \cline{2-3}
 & DR+UP (True) & $7586.8\pm{3435.7}$  \\ \cline{2-3}
 & DR+UP+SI & $2153.9\pm{3741.0}$\\ \cline{2-3}
 & \textbf{DR+UP+Encoding (BO)} & $\mathbf{7303.9\pm{3574.2}}$ \\ \cline{2-3}
 & DR+UP+Encoding(True) & $7474.0\pm{3454.7}$ \\ \hline
 
 \multirow{6}{*}{\textit{halfcheetah}} & No DR & $3213.6\pm{2917.9}$  \\ \cline{2-3}
 & DR only & $7594.2\pm{2571.1}$\\ \cline{2-3}
 & DR+UP (True)& $8956.7\pm{2279.9}$  \\ \cline{2-3}
  & DR+UP+SI & $5679.7\pm{2994.4}$\\ \cline{2-3}
 & \textbf{DR+UP+Encoding (BO)} & $\mathbf{8051.5\pm{2646.6}}$ \\ \cline{2-3}
 & DR+UP+Encoding(True) & $8194.5\pm{2385.5}$ \\ 
\hline
\end{tabular}
\label{tab:test}
\end{center}
\end{table}

Fig.~\ref{fig:compare_methods} and Tab.~\ref{tab:test} display the test performances of different methods on randomly sampled system parameters within the same distributions of DR, as the step (d) in Fig.~\ref{fig:overview}. We show the episode reward distributions of different methods as a violin plot in Fig.~\ref{fig:compare_methods}. Policies for each method are tested on 10 randomly sampled environment dynamics for 10 episodes each, as a total of 100 episodes for each method. We can see that although the method without DR can achieve good rewards in training since there is no randomized dynamics but a fixed one, in test cases it has a heavy allocation for both the head and tail on the performance distribution. The reason is that the lack of randomized dynamics show some unseen cases incapable of being handled well by the policy. The method with only DR also does not work well because of the hardness in training an optimal policy in randomized environments, which is testified to be an ill-posed problem. For both environments, the heavy tails in no-DR method are greatly alleviated for methods with UP, due to its awareness of system dynamics. However, the true dynamics parameters of the testing environment are usually not accessible, which requires a SI module to configure. Our experiments show that the SI process with a stack of 5 frames of historical transitions (see similar settings in~\cite{yu2017preparing}) is not capable of providing an accurate estimation of the system dynamics,  which severely degrades the performances for UP methods. For both environments, our proposed method with embedding SI using BO process shows advantageous performances over other methods, even as good as the one with true system parameters.

\section{CONCLUSIONS and DISCUSSIONS}
We propose to optimize the dynamics embedding rather than directly configuring the system parameters from historical transitions, and demonstrate its advantageous performances over standard DR and UP methods for a general domain transfer setting, with some primary tests on both a low-dimensional and a high-dimensional simulated environments. The deficiency for DR policies of not being able to achieve optimal actions and the difficulties within normal SI methods for directly configuring the system parameters are revealed. Our future work involves extending the current framework to more complex robot learning tasks, as well as its application on sim-to-real transfer problem, which is a subset of the domain transfer in general. The difficulty of UP learning due to the expanded input spaces with high-dimensional system parameters are not exposed in current experiments, therefore some higher-dimensional tasks will be investigated as well.




\section*{APPENDIX}
\subsection{Theoretical Proofs}
\label{appsec:proof}
\begin{customlemma}{2}
Given the same dataset distribution $\rho^s(s,a,s')=\rho^t(s,a,s')$, the difference between the estimated dynamics distribution $p^s$ and the true distribution $p^t$ can be characterized as:
\begin{align}
&\mathbb{D}_\text{KL}[p^s(\theta|s,a,s')||p^t(\theta|s,a,s')] \\
    &=\mathbb{D}_\text{KL}[\rho^s(s,a,\theta,s')||\rho^t(s,a,\theta,s')]
\end{align}
\end{customlemma}
\begin{proof}
\begin{align}
\text{RHS}&=\mathbb{D}_\text{KL}[\rho^s(s,a,\theta,s')||\rho^t(s,a,\theta,s')]\\
&-\mathbb{D}_\text{KL}[\rho^s(s,a,s')||\rho^t(s,a,s')] \\
&=\int_{\mathcal{S}\times\mathcal{A}\times\Theta\times\mathcal{S}}\rho^s(s,a,\theta,s')\log(\frac{\rho^s(s,a,\theta,s')}{\rho^t(s,a,\theta,s')} \\ &\times\frac{\rho^t(s,a,s')}{\rho^s(s,a,s')})\text{d}s\text{d}a\text{d}\theta\text{d}s' \\
&=\int_{\mathcal{S}\times\mathcal{A}\times\Theta\times\mathcal{S}}\rho^s(s,a,\theta,s')\log\frac{p^s(\theta|s,a,s')}{p^t(\theta|s,a,s')}\text{d}s\text{d}a\text{d}\theta\text{d}s' \\
&=\int_{\Theta}p^s(\theta|s,a,s')\log\frac{p^s(\theta|s,a,s')}{p^t(\theta|s,a,s')}\text{d}\theta \\
&=\mathbb{D}_\text{KL}[p^s(\theta|s,a,s')||p^t(\theta|s,a,s')] = \text{LHS}.
\end{align}
\end{proof}

\begin{customlemma}{4} The distance of the distribution from the forward dynamics prediction and the true distribution can be formulated with the KL-divergence, it thus follows:
\begin{align}

\mathbb{D}_\text{KL}[f^s||f^t]\approx \mathbb{D}_\text{KL}[\rho^s||\rho^t] -\mathbb{D}_\text{KL}[\rho^s(s,a,\theta)||\rho^t(s,a,\theta)]
\end{align}
where $f^{s,t}$ are shorten for $f^{s,t}(s'|s,a, \theta)$, and $\rho^{s,t}$ are shorten for $\rho^{s,t}(s,a,\theta,s')$.
\end{customlemma}

\begin{proof}
Similar as the proof of Lemma~\ref{the:1}, we have,
\begin{align}
    \text{RHS}&=\mathbb{D}_\text{KL}[\rho^s(s,a,\theta,s')||\rho^t(s,a,\theta,s')] \\
    &- \mathbb{D}_\text{KL}[\rho^s(s,a,\theta)||\rho^t(s,a,\theta)] \\
&=\int_{\mathcal{S}\times\mathcal{A}\times\Theta\times\mathcal{S}}\rho^s(s,a,\theta,s')\log(\frac{\rho^s(s,a,\theta,s')}{\rho^t(s,a,\theta,s')} \\
&\times\frac{\rho^t(s,a,\theta)}{\rho^s(s,a,\theta)})\text{d}s\text{d}a\text{d}\theta\text{d}s' \\
&\approx\int_{\mathcal{S}\times\mathcal{A}\times\Theta\times\mathcal{S}}\rho^s(s,a,\theta,s')\log\frac{f^s(s'|s,a,\theta)}{f^t(s'|s,a,\theta)}\text{d}s\text{d}a\text{d}\theta\text{d}s' \\
&=\int_{\Theta}f^s(s'|s,a,\theta)\log\frac{f^s(s'|s,a,\theta)}{f^t(s'|s,a,\theta)}\text{d}s' \\
&=\mathbb{D}_\text{KL}[f^s(s'|s,a,\theta)||f^t(s'|s,a,\theta)] = \text{LHS}
\end{align}
\end{proof}

\begin{customthm}{5}
The optimization of forward dynamics prediction is increasing the lower bound of the optimization objective for improving the estimated dynamics, i.e.,

\begin{equation}
    \mathbb{D}_\text{KL}[p^s||p^t]\geq  \mathbb{D}_\text{KL}[f^s||f^t]
\end{equation}
where $p^{s,t}$ are $p^{s,t}(\theta|s,a,s')$ and $f^{s,t}$ are $f^{s,t}(s'|s,a,\theta)$.
\end{customthm}

\begin{proof}

With Lemma~\ref{the:1} and \ref{the:2}, we have,
\begin{align}
    \text{LHS}&=\mathbb{D}_\text{KL}[\rho^s(s,a,\theta,s')||\rho^t(s,a,\theta,s')]\\
    &=\mathbb{D}_\text{KL}[\rho^s(s,a,\theta,s')||\rho^t(s,a,\theta,s')] \\
    &- \mathbb{D}_\text{KL}[\rho^s(s,a,\theta)||\rho^t(s,a,\theta)] + \mathbb{D}_\text{KL}[\rho^s(s,a,\theta)||\rho^t(s,a,\theta)]\\
    &\approx \mathbb{D}_\text{KL}[f^s(s'|s,a,\theta)||f^t(s'|s,a,\theta)] \\
    &+ \mathbb{D}_\text{KL}[\rho^s(s,a,\theta)||\rho^t(s,a,\theta)]\\
    &\geq \mathbb{D}_\text{KL}[f^s(s'|s,a,\theta)||f^t(s'|s,a,\theta)]
\end{align}
\end{proof}
The last line of above proof is due to that $\mathbb{D}_\text{KL}[\rho^s(s,a,\theta)||\rho^t(s,a,\theta)]$ is non-negative. 

\section{Randomized Parameters}
\label{appsec:rand}
\begin{table}[h!]
\begin{center}
\captionof{table}{Randomized dynamics parameters (5-dimensional) for \textit{InvertedDoublePendulum-v2}}
\begin{tabular}{ |c|c|c| } 
\hline
\textbf{Variable} & \textbf{Range} \\
\hline
damping & [0.02, 0.3] \\ \hline
gravity & [8.5, 11.0] \\ \hline
length1& [0.3, 0.9] \\ \hline
length2 & [0.3, 0.9] \\ \hline
density & [0.5, 1.5] \\ 
\hline
\end{tabular}
\label{tab:pendulum}
\end{center}
\end{table}


\begin{table}[h!]
\begin{center}
\captionof{table}{Randomized dynamics parameters (13-dimensional) for \textit{HalfCheetah-v3}}
\begin{tabular}{ |c|c|c| } 
\hline
\textbf{Variable} & \textbf{Range} \\
\hline
gravity & [5.5, 14.0] \\ \hline
bthigh damping & [3.0, 9.0] \\ \hline
bshin damping & [1.5, 7.5] \\ \hline
bfoot damping & [1.0, 5.0] \\ \hline
fthigh damping & [1.5, 7.5] \\ \hline
fshin damping & [1.0, 5.0] \\ \hline
ffoot damping & [0.2, 2.8] \\ \hline
bthigh stiffness & [100, 380] \\ \hline
bshin stiffness & [20, 340] \\ \hline
bfoot stiffness & [10, 230] \\ \hline
fthigh stiffness & [20, 340] \\ \hline
fshin stiffness & [20, 220] \\ \hline
ffoot stiffness & [10, 110] \\
\hline
\end{tabular}
\label{tab:halfcheetah}
\end{center}
\end{table}

\bibliography{ref}
\bibliographystyle{IEEEtran}

\end{document}